\documentclass[11pt]{article}

\usepackage[utf8]{inputenc} % allow utf-8 input
\usepackage[T1]{fontenc}    % use 8-bit T1 fonts
\usepackage{lmodern}
\usepackage{url}            % simple URL typesetting
\usepackage{booktabs}       % professional-quality tables
\usepackage{amsfonts}       % blackboard math symbols
\usepackage{nicefrac}       % compact symbols for 1/2, etc.
\usepackage{microtype}      % microtypography
\usepackage{float}
\usepackage{enumerate}

\usepackage{tikz}
\usetikzlibrary{positioning}

\usepackage{mathrsfs}
\usepackage{amsmath}
\usepackage{amssymb}
\usepackage{graphicx}
\usepackage{url}
\PassOptionsToPackage{x11names}{xcolor}
\usepackage{xcolor}
\PassOptionsToPackage{algo2e,ruled}{algorithm2e}
\usepackage{algorithm2e}
\usepackage{jmlrutils}

\usepackage{hyperref}
\usepackage{nameref}

\newtheorem{observation}{Observation}

\usepackage{import}

\usepackage{authblk}%\usepackage{xr}
\usepackage{pdfpages} % in preample

\begin{document}
	\setcounter{page}{1}
	\title{Explicit regularization and implicit bias in  deep
		network classifiers trained with the square loss}
		\author{\large Tomaso Poggio and Qianli Liao}
	\date{\today}
	
	\maketitle
	
	\begin{abstract}
		
          Deep ReLU networks trained  with the square loss have been observed to perform well in
          classification tasks. We provide here a theoretical
          justification based on analysis of the associated gradient
          flow. We show that convergence to a
          solution with the absolute minimum norm is expected when
          normalization techniques such as Batch
          Normalization (BN) or Weight
          Normalization (WN) are used together with Weight Decay
          (WD). The main property of the minimizers
          that bounds their expected error is the norm: we prove that
          among all the close-to-interpolating solutions, the ones associated
          with smaller Frobenius norms of the unnormalized weight matrices have
          better margin and better bounds on the expected
          classification error. With BN but in the absence of WD, the dynamical
          system is singular. Implicit dynamical regularization --
          that is zero-initial conditions biasing the dynamics towards
          high margin solutions -- is also possible in the no-BN and
          no-WD case.  The theory yields several predictions,
          including the role of BN and weight decay, aspects of Papyan,
          Han and Donoho's Neural Collapse and the constraints induced
          by BN on the network weights.

\end{abstract}

\section{Introduction}

In the case of exponential-type loss functions a mechanism of
complexity control underlying generalization was identified in the asymptotic margin maximization effect of
minimizing exponential-type loss functions
\cite{2019arXiv190507325S,DBLP:journals/corr/abs-1906-05890,
  theory_III}. However, this mechanism

\begin{itemize}
\item cannot explain the good
empirical results that have been recently demostrated using the square loss\cite{hui2020evaluation};
\item cannot explain the empirical evidence that convergence for cross-entropy loss
  minimization depends on initialization.
\end{itemize}

This puzzle motivates our focus in this paper on the square loss.
More details and sketches of the proofs can be found in
\cite{PoggioLiaoArxiv2020}.

Here we {\it assume} commonly used GD-based normalization algorithms
such such as BN\cite{ioffe2015batch} (or WN\cite{SalDied16}) together
with weight decay (WD), since they appear to be essential for
reliably\cite{DBLP:journals/corr/abs-1812-03981} training deep
networks (and were used by \cite{hui2020evaluation}. We also consider,
however, the case in which neither BN nor WD are used\footnote{This
  was the main concern of the first version of
  \cite{PoggioLiaoArxiv2020}.} showing that a dynamic ``implicit
regularization'' effect for classification is still possible with convergence strongly
depending on initial conditions.

\subsection{Notation}

We define a deep network with $L$ layers with the usual
coordinate-wise scalar activation functions
$\sigma(z):\quad \mathbf{R} \to \mathbf{R}$ as the set of functions
$g(W;x) = (W_L \sigma (W_{L-1} \cdots \sigma (W_1 x)))$, where the
input is $x \in \mathbf{R}^d$, the weights are given by the matrices
$W_k$, one per layer, with matching dimensions. We sometime use the
symbol $W$ as a shorthand for the set of $W_k$ matrices
$k=1,\cdots,L$. There are no bias terms: the bias is instantiated in
the input layer by one of the input dimensions being a constant.  The
activation nonlinearity is a ReLU, given by
$\sigma (x) = x_+ = max(0, x)$ . Furthermore,

\begin{itemize}

\item we define $g(x)=\rho f(x)$ with $\rho$ defined as the product
  of the Frobenius norms of the weight matrices of the $L$ layers of
  the network and $f$ as the corresponding network with normalized
  weight matrices $V_k$ (because the ReLU is homogeneous
  \cite{theory_III});
\item in the following we use the notation $f_n$ meaning $f(x_n)$,
  that is the ouput of the normalized
  network for the input $x_n$;
\item we assume $||x||=1$;

\item {\it separability} is defined as correct classification for all
  training data, that is $y_n f_n>0,\quad \forall n$. We call {\it
    average separability} when $\sum y_nf_n>0$.

\end{itemize}

\subsection{Regression and classification}

In our analysis of the square loss, we need to explain when and why
{\it regression works well for classification}, since the training minimizes
square loss but we are interested in good performance in
classification (for simplicity we consider here binary
classification).  Unlike the case of linear networks we expect several
global zero square loss minima corresponding to interpolating
solutions (in general degenerate, see \cite{PoggioCooper} and
reference therein). Although all interpolating solutions are optimal
solutions of the regression problem, they will in general have
different margins and thus different expected classification
performance. In other words, zero square loss does not imply by itself
neither large margin nor good expected classification. Notice that if
$g$ is a zero loss solution of the regression problem, then
$g(x_n)=y_n, \quad \forall n$. This is equivalent to $\rho f_n=y_n$
where $f_n$ is the {\it margin} for $x_n$. Thus the norm $\rho$ of a
minimizer is inversely related to its average margin. In fact, for an
exact zero loss solution of the regression problem, the margin is the
same for all training data $x_n$ and it is equal to
$\frac{1}{\rho_{eq}}$. Starting from small initialization, GD will
explore critical points with $\rho$ growing from zero, as we will
show. Thus interpolating solutions with small norm $\rho_{eq}$
(corresponding to the best margin) may be found before large
$\rho_{eq}$ solutions which have worse margin. If the weight decay
parameter is non-zero and large enough, there is independence from initial
conditions. Otherwise, a near-zero initialization is required, as in the
case of linear networks, though the reason is quite different and is
due to an implicit bias in the dynamics of GD.

\section{Dynamics and Generalization}
\label{Lagrange}

Our {\it key assumption} is that the main property of batch
normalization (BN) and weight normalization (WN) -- the normalization
of the weight matrices -- can be captured by 
the  gradient flow on a loss function modified by adding Lagrange
multipliers.

Gradient descent on a modified square loss
\begin{equation}
\mathcal{L}=\sum_n(\rho f_n-y_n)^2 +
\nu \sum_k ||V_k||^2
\label{LagrangeMultiplierEquation}
\end{equation} 
\noindent with $ ||V_k||^2=1$
is in fact exactly equivalent to ``Weight Normalization'', as proved in \cite{theory_III}, for deep
networks.

This  dynamics can be written as
        $\dot{\rho_k} =V_k^T \dot{W}_k$ and
        $\dot{V_k}= \rho S \dot{W}_k$ with $S=I-V_k V_k^T$. This shows
        that if $W_k= \rho_k V_k$ then
        $\dot{V_k}= \frac{1}{\rho_k} \dot{W_k}$ as mentioned in
        \cite{DBLP:journals/corr/abs-1812-03981}.  The condition
        $||V_k||^2=1$  yields, as shown in \cite{PoggioLiaoArxiv2020},  $\nu= - \sum_n(\rho^2
          f_n^2- \rho y_n f_n)$.
Here we assume  that the BN module is used in all layers apart the
last one, that is we assume
        $\rho_k= 1, \quad \forall k<L$ and $\rho_L=\rho$ where $L$ is
        the number of layers\footnote{It is important to observe here that
        batch normalization -- unlike Weight Normalization --  leads
        not only to
          normalization of the weight matrices but also to
          normalization of each row of the weight matrices
          \cite{theory_III} because it normalizes separately the
          activity of each unit
          $i$ and thus -- indirectly -- the $W_{i,j}$ for each $i$ separately. This 
          implies that each
          row $i$ in $(V_k)_{i,j}$ is normalized independently and thus the whole
          matrix $V_k$ is normalized (assuming the normalization of
          each row is the same $1$ for all rows). The equations in the
          main text involving $V_k$ can be read in this way, that is
          restricted to each row.}.

          As we will show, the dynamical system associated with the
          gradient flow of the Lagrangian Equation
          \ref{LagrangeMultiplierEquation} is ``singular'', in the
          sense that normalization is not guaranteed at the critical
          points. Regularization is needed, and in fact it is common to
          use in gradient descent not only batch normalization but
          also weight decay.  Weight decay consists of a regularization
          term $\lambda ||W_k||^2$ added to the Lagrangian yielding

          \begin{equation}
\mathcal{L}=\sum_n(\rho f_n-y_n)^2 +
\nu \sum_k ||V_k||^2+ \lambda \rho^2.
\label{LagrangianNormalizationRegularization}
\end{equation} 

The associate gradient flow is then the following dynamical system

\begin{equation}
	\dot{\rho}=  -2
        [\sum_n \rho ( f_n)^2- \sum_n f_n y_n] -2 \lambda \rho
        \label{DynSys1}
	\end{equation}

\begin{equation}
	\dot{V_k}= 2 \rho
        \sum_n[(\rho f_n-y_n)  ( V_k f_n - \frac{\partial f_n}{\partial
          V_k})]
                \label{DynSys2}
	\end{equation}

        \noindent where the critical points
        $\dot{\rho}=0, \quad \dot{V_k}=0$ are singulat for $\lambda=0$
        but are not singular for any
        arbitrarily small
        $\lambda>0$\footnote{For $\lambda=0$ the zero loss critical point is
          pathological, since $\dot{V_k}=0$ even when $ ( V_k f_n - \frac{\partial f_n}{\partial
          V_k})$ implying that an un-normalized interpolating solution
        satisfies the equilibrium equations. Numerical simulations
        show that even for  linear degenerate networks convergence
        is independent of initial conditions only if  $\lambda>0$.}.
 In particular, normalization is then effective at
        $\rho_{eq}$ unlike in the $\lambda=0$ case. As a side remark, SGD,
        as opposed to gradient flow, may help (especially
        with label noise) to counter to some extent
        the singularity of the $\lambda=0$ case, even without weight
        decay, because of the associated random fluctuations around
        the pathological critical point.

The equilibrium
       value at  $\dot{\rho_k}=0$   is

	\begin{equation}
          \rho_{eq}=\frac{\sum_n y_nf_n} {\lambda +\sum_n f^2_n}.
          \label{lambdaweightdecay}
	\end{equation}

       {\it Observe that $\dot{\rho} >0$ if $\rho$ is smaller than
        $\rho_{eq}$ and if average separability holds.} Recall also
        that zero loss ``global'' minima (in fact arbitrarily close to
        zero for small but positive $\lambda$) are expected {\it to exist and be
        degenerate} \cite{PoggioCooper}.

        If we assume that the loss (with the constraint
        $||V_k||=1$) is a continuous function of the $V_k$, then {\it there
        will be at least one minimum of $\mathcal{L}$ at any fixed $\rho$, because the
        domain $V_k$ is compact}. This means that for each $\rho$ there
        is at least a critical point of the gradient flow of $V_k$, implying that for {\it
          each critical $\rho$ for which $\dot{\rho}=0$, there is at
          least one 
          critical point of the dynamical system in $\rho$ and $V_k$}.

Around $\dot{V_k}=0$  we have
\begin{equation} 
 \sum_n  (\rho f_n - y_n)  {\frac{\partial f_n}{\partial V_k }}
 =\sum_n  (\rho f_n - y_n) (V^{eq}_k f_n),
\label{V-GD3}
\end{equation}

\noindent where the terms $ (\rho f_n - y_n)$ will be generically
different from zero if $\lambda>0$.

The conclusions of this analysis can be summarized in 

\begin{observation}
  Assuming {\it average separability}, and gradient flow starting from
  small norm norm 
  $\rho(0)=\epsilon$ >0, $\rho(t)$ grows monotonically until a minimum is reached at
  which $\rho_{eq}=\frac{\sum_n y_nf_n} {\lambda +\sum_n f^2_n}.$ This
  dynamics is expected  even in the limit of $\lambda=0$, which corresponds to
  exact interpolation of all the training data at a singular critical point.
\end{observation}

  and

\begin{observation}
  Minimizers with small $\rho_{eq}$ correspond to large average margin
  $\sum y_n f_n$. In particular, suppose that the gradient flow converges to a
  $\rho_{eq}$ and $V_k^{eq}$ which correspond to zero square
  loss. Among all such minimizers the one with the  smallest
  $\rho_{eq}$ (typically found first during the GD dynamics when
  $\rho$ increases from $\rho=0$),
  corresponds to the (absolute) minimum norm -- and maximum margin --
  solutions.  
\end{observation}

In general, there may be several critical points of the $V_k$ for the
same $\rho_{eq}$ and they are typically degenerate (see references in
\cite{PoggioCooper2020}) with dimensionality $W-N$, where $W$ is the
number of weights in the network may be degenerate. All of them will
correspond to the same norm and all will have the same margin for all
of the training points.

%\section{Dynamics}

Since
usually the maximum output of a multilayer
network   is $<<1$, the first critical point for increasing $\rho$ will
be when $\rho$ becomes  large enough to allow the following equation to
have solutions 

\begin{equation}
\sum_n  y_n f_n=\rho (\lambda +\sum_n f^2_n).
\end{equation}

If gradient flow starts from very small $\rho$ and there is {\it
  average separability}, $\rho$ increases monotonically until such a
minimum is found. If $\rho$ is
  large, then $\dot{\rho} <0$ and $\rho$ will decrease until a minimum
  is found.

For large $\rho$ and very small $\lambda$, we expect many  solutions  under GD\footnote{It is interesting to recall
\cite{PoggioCooper} that for SGD -- unlike GD -- the algorithm will
stop only when $\ell_n=0 \quad \forall n$, which is the global minimum
and corresponds to perfect interpolation. For the other critical
points for which GD will stop, SGD will never stop but just fluctuate
around the critical point.}. 
The emerging picture is a landscape in
which there are no zero-loss  minima for $\rho < \rho_{min}$. With increasing
$\rho$ from $\rho=0$ there will be   zero square-loss degenerate
minima
with the minimizer representing an almost interpolating solution (for $\lambda>0$). We expect,
however, that depending on the value of $\lambda$, there is a bias towards minimum $\rho_{eq}$
even for large $\rho$ initializations and certainly for small intializations.

\begin{figure}
  \centering
  \includegraphics[width=1.0\textwidth]{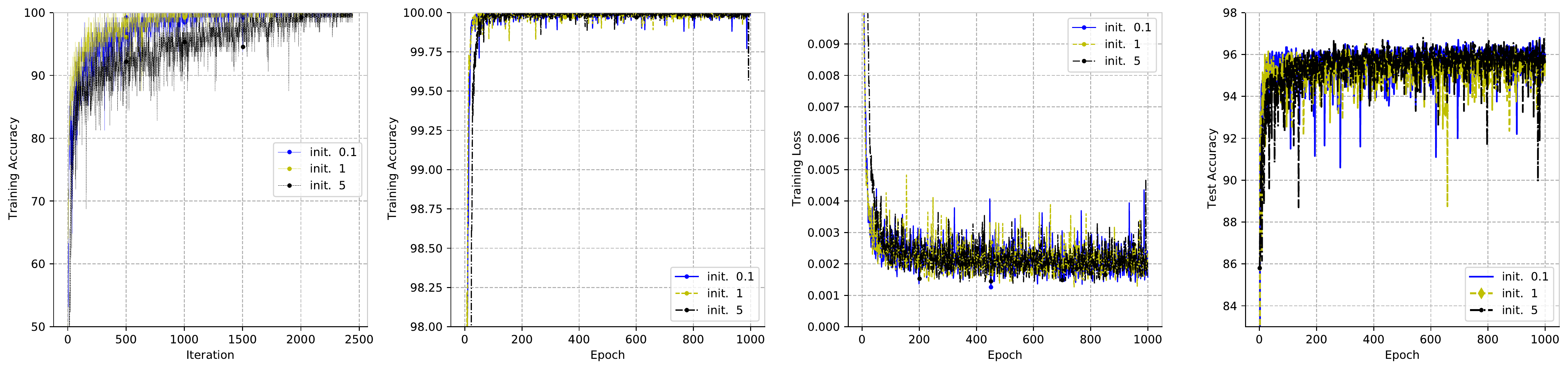}             
  \caption {\it  \textbf{ConvNet with Batch Normalization and Weight Decay} Binary classification on two classes from CIFAR-10, trained with MSE loss. The model is a very simple network with 4 layers of fully-connected Layers. ReLU nonlinearity is used. Batch normalization is used. The weight matrices of all layers are initialized with zero-mean normal distribution, scaled by a constant such that the Frobenius norm of each matrix is 5. We use weight decay of 0.01. We run SGD with batch size 128, constant learning rate 0.1 and momentum 0.9 for 1000 epochs. No weight decay.  No data augmentation.    Every input to the network is scaled such that it has Frobenius norm 1.  }                      
  \label{appendix:fig:conv4_BN_wd_0.01:training_val}
\end{figure}

\begin{figure}
  \centering
  \includegraphics[width=1.0\textwidth]{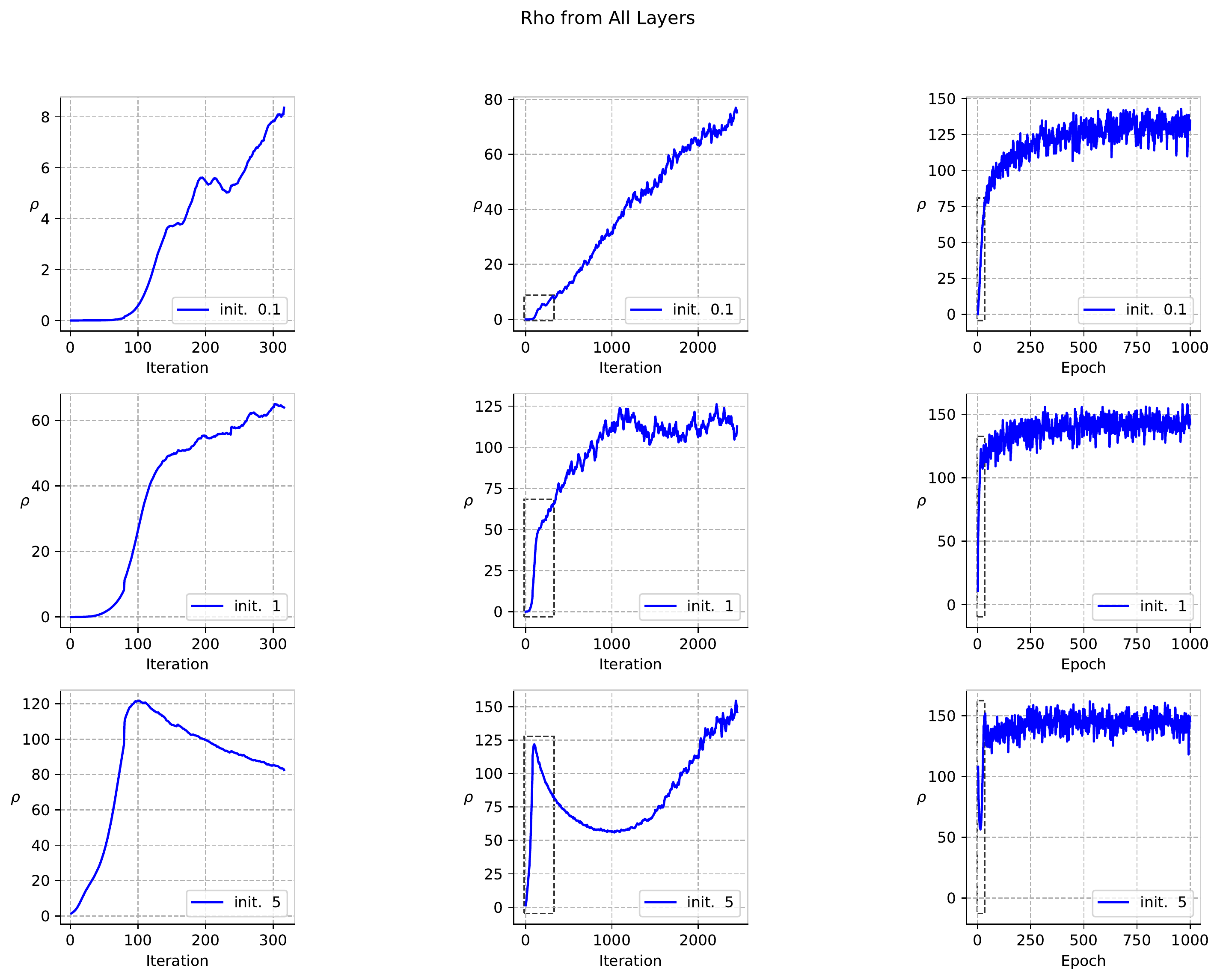}
  \caption {\it  \textbf{ConvNet with Batch Normalization and Weight Decay} Dynamics of $\rho$ from experiments in Figure \ref{appendix:fig:conv4_BN_wd_0.01:training_val}. First row: small initialization (0.1).  Second row: medium initialization (1).  Third row:  large initialization (5). A dashed rectangle denotes the previous subplot's domain and range in the new subplot.      }       
  \label{appendix:fig:conv4_BN_wd_0.01:rho}
\end{figure}

\begin{figure}
  \centering 
  \includegraphics[width=1.0\textwidth]{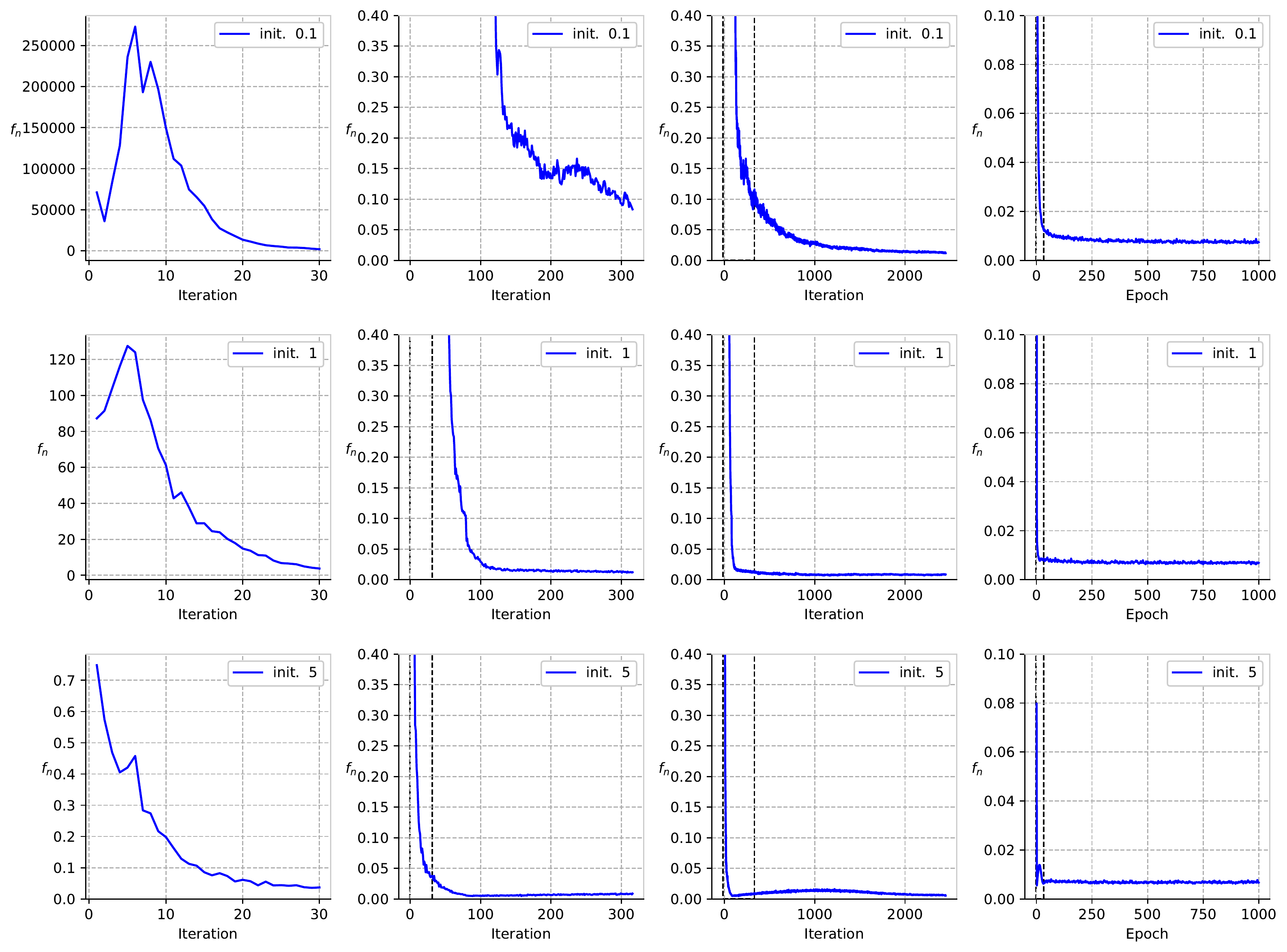} 
  \caption {\it  \textbf{ConvNet with Batch Normalization and Weight Decay} Dynamics of the average of $|f_n|$ from experiments in Figure \ref{appendix:fig:conv4_BN_wd_0.01:training_val}. First row: small initialization (0.1).  Second row: medium initialization (1).  Third row:  large initialization (5). A dashed rectangle denotes the previous subplot's domain and range in the new subplot.      }              
  \label{appendix:fig:conv4_BN_wd_0.01:mean_abs_fn}
\end{figure}

\begin{figure}
  \centering
  \includegraphics[width=1.0\textwidth]{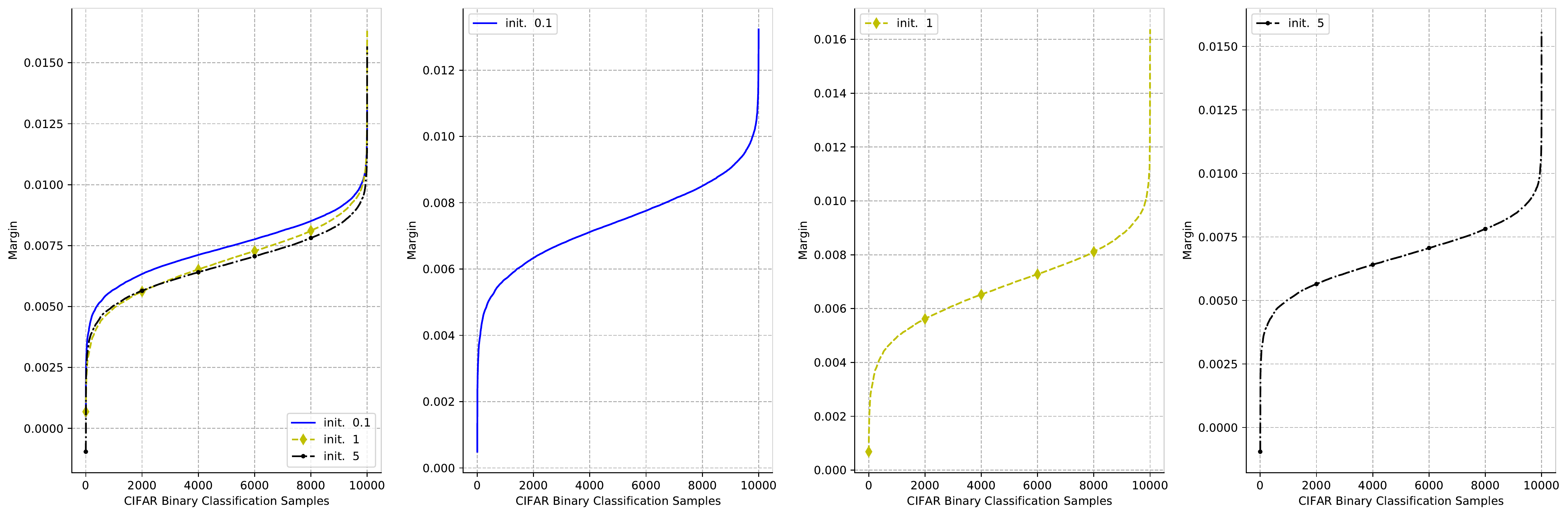} 
  \caption {\it  \textbf{ConvNet with Batch Normalization and Weight Decay} Margin of all training samples.}  \label{appendix:fig:conv4_BN_wd_0.01:margin}    
\end{figure}

All these observations are also supported by our numerical
experiments.  Figure \ref{appendix:fig:conv4_BN_wd_0.01:training_val},
\ref{appendix:fig:conv4_BN_wd_0.01:rho},
\ref{appendix:fig:conv4_BN_wd_0.01:mean_abs_fn},
%\ref{appendix:fig:conv4_BN_wd_0.01:ratio}
and
\ref{appendix:fig:conv4_BN_wd_0.01:margin} show the case of gradient
descent with batch normalization and weight decay, which corresponds
to a well-posed dynamical system for gradient flow; the other figures
show the same networks and data with BN without WD and without both BN
and WD. As predicted by the analysis, the case of BN+WD is the most
well-behaved, whereas the others strongly depend on initial
conditions.

%
%\begin{figure}
%  \centering
%  \includegraphics[width=1.0\textwidth]{figures/BN_weight_decay_0.01/rho_conv.pdf}
%  \caption {\it  \textbf{ConvNet with Batch Normalization and Weight Decay} Dynamics of $\rho$ of only convolution layers from experiments in Figure \ref{appendix:fig:conv4_BN_wd_0.01:training_val}. First row: small initialization (0.1).  Second row: medium initialization (1).  Third row:  large initialization (5). A dashed rectangle denotes the previous subplot's domain and range in the new subplot.      }          
%  \label{appendix:fig:conv4_BN_wd_0.01:rho}
%\end{figure}

%\begin{figure}
%  \centering
%  \includegraphics[width=1.0\textwidth]{figures/BN_weight_decay_0.01/ratio.pdf}
%  \caption {\it  \textbf{ConvNet with Batch Normalization and Weight Decay} Ratio of $\frac{\sum y_nf_n}{\sum f_n^2} $ and $\rho$.   First row: small initialization (0.1).  Second row: medium initialization (1).  Third row:  large initialization (5).  
%}  \label{appendix:fig:conv4_BN_wd_0.01:ratio}
%\end{figure}

%\section{Generalization in Deep Networks}

To show that $\rho$ indeed controls the expected error we use
classical bounds that lead  to the
following theorem

\begin{observation}
  \label{boundtp}
 With probability $1-\delta$
\begin{equation}
L(f) \leq  c_1 \rho  \mathbb{R}_N(\tilde{\mathbb{F}})+ c_2 \epsilon (N, \delta)
\end{equation}
\noindent where $c_1, c_2$ are constants that reflect the Lipschitz constant of  the loss
function ( for the square loss this requires a bound on $f(x)$) and
the architecture of the network. The Rademacher average
$\mathbb{R}_N(\tilde{\mathbb{F}})$ depends on the normalized network
architecture and $N$. Thus for the same network and the same data, the upper bound for the expected
error of the minimizer is smaller for smaller $\rho$.
\vspace{0.1in}
\end{observation}

The theorem proves the conjecture in \cite{Foundations} that for deep
networks, as for kernel machines, {\it minimum norm interpolating
  solutions are the most stable}.

\begin{figure}
  \centering
  \includegraphics[width=1.0\textwidth]{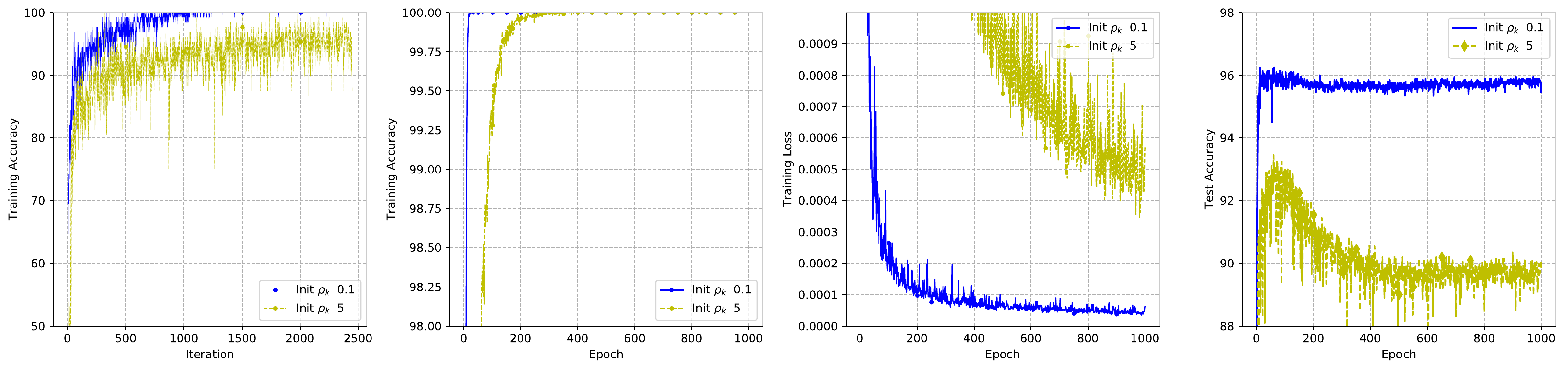}   
  \caption {\it \textbf{ConvNet with Batch Normalization but no Weight Decay}  Binary classification on two classes from CIFAR-10,
    trained with MSE loss. The model is a very simple network with 4
    layers of convolutions. ReLU nonlinearity is used. Batch
    normalization is used without parameters (affine=False in
    PyTorch). The weight matrices of all layers are initialized with
    zero-mean normal distribution, scaled by a constant such that the
    Frobenius norm of each matrix is either 0.1 or 5. We run SGD with
    batch size 128, constant learning rate 0.01 and momentum 0.9 for
    1000 epochs. No data augmentation.   Every input
    to the network is scaled such that it has Frobenius norm 1. This
    is a single run but it is typical for the parameter values we used. }          
  \label{fig:convBN:training_val}
\end{figure}

\begin{figure}
  \centering
  \includegraphics[width=1.0\textwidth]{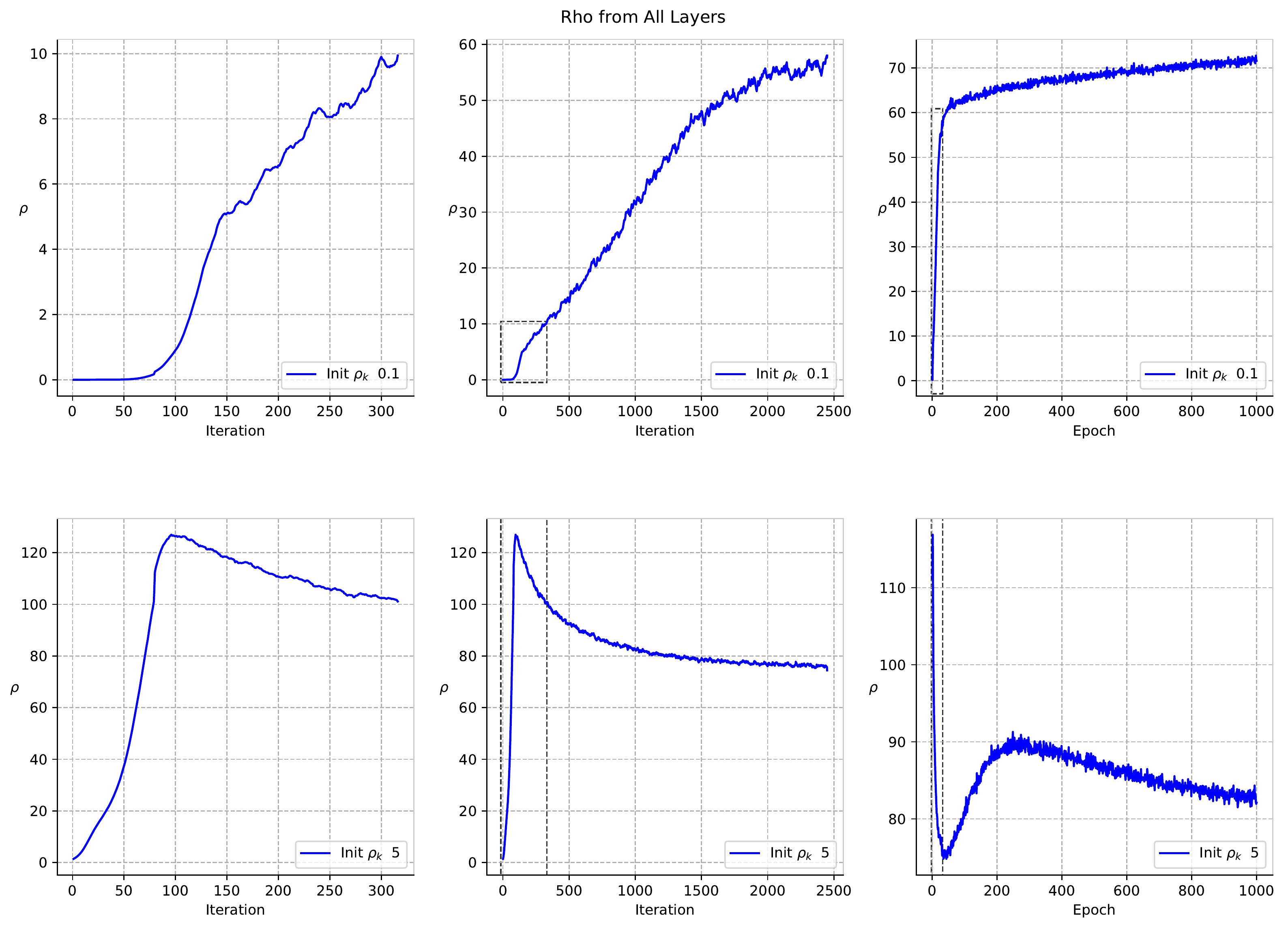}
  \caption {\it \textbf{ConvNet with Batch Normalization but no Weight Decay}. Dynamics of $\rho$ from experiments in Figure
    \ref{fig:convBN:training_val}. Top row: small initialization
    (0.1).  Bottom row: large initialization (5). The plot starts with
    $\rho(0)=0$ despite an initialization of $\rho_k=0$ because the the scaling factor of BN starts from $0$. A dashed rectangle denotes the previous subplot's domain and range in the new subplot.      }      
  \label{fig:convBN:rho}
\end{figure}

\begin{figure}
  \centering
  \includegraphics[width=1.0\textwidth]{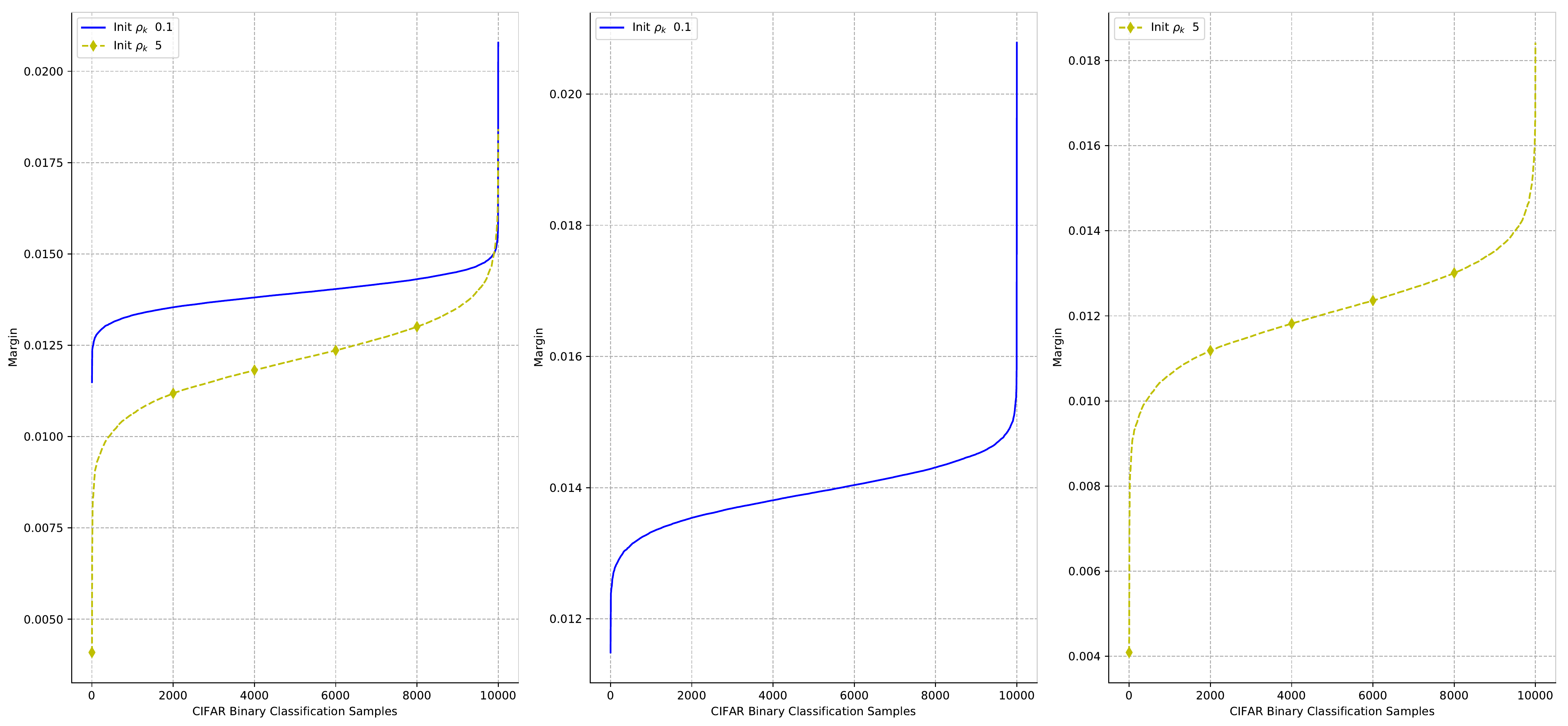}
  \caption {\it \textbf{ConvNet with Batch Normalization but no Weight Decay}. Margin of all training samples (see previous
    figures). If the solution were to correspond to exactly zero
    square loss, the margin distribution would be an horizontal line. }  \label{fig:convBN:margin}
\end{figure}

\begin{figure}
  \centering
  \includegraphics[width=1.0\textwidth]{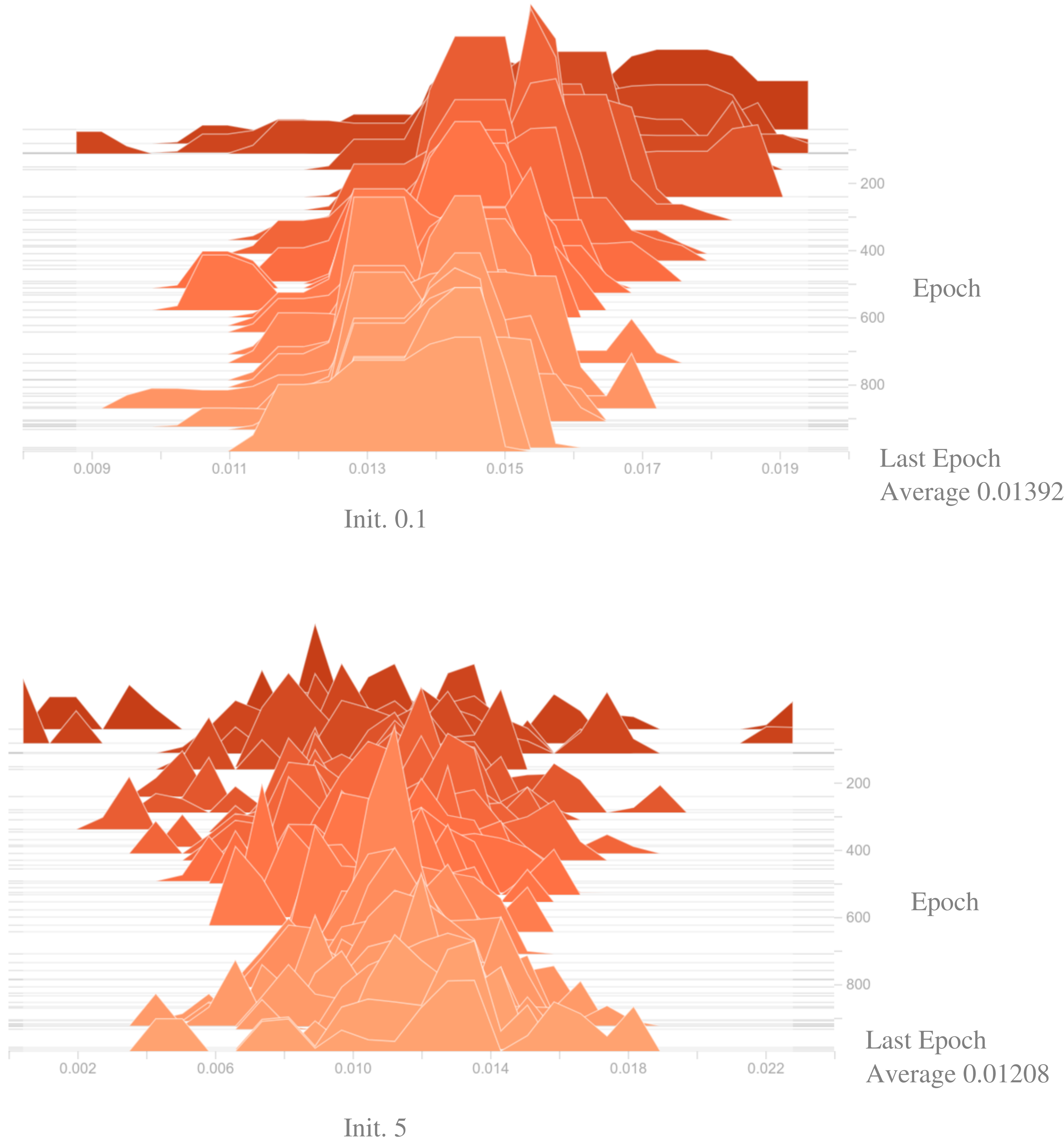}
  \caption {\it \textbf{ConvNet with Batch Normalization but no Weight Decay}: Histogram of $|f_n|$ over time. Top figure: initial $\rho_k=0.1$.  Bottom figure: initial $\rho_k=5$.}  \label{fig:convBN:f_n}      
\end{figure}

\subsection{Predictions}
 \begin{itemize}

 \item In a recent paper Papyan, Han and Donoho\cite{Papyan24652}
   described four empirical properties of the terminal phase of
   training (TPT) deep networks, using the cross-entropy loss
   function. TPT begins at the epoch where training error first
   vanishes. During TPT, the training error stays effectively zero,
   while training loss is pushed toward zero. Direct empirical
   measurements expose an inductive bias they call neural collapse
   (NC), involving four interconnected phenomena. (NC1) Cross-example
   within-class variability of last-layer training activations
   collapses to zero, as the individual activations themselves
   collapse to their class means. (NC2) The class means collapse to
   the vertices of a simplex equiangular tight frame (ETF). (NC3) Up
   to rescaling, the last-layer classifiers collapse to the class
   means or in other words, to the simplex ETF (i.e., to a self-dual
   configuration). (NC4) For a given activation, the
   classifier{\textquoteright}s decision collapses to simply choosing
   whichever class has the closest train class mean (i.e., the nearest
   class center [NCC] decision rule).  We show in \cite{PoggioLiaoArxiv2020} that these {\it properties of the Neural
     Collapse\cite{Papyan24652} seem to be predicted} by the theory of
   this paper for the global (that is, close-to-zero square-loss)
   minima, irrespectively of the value of $\rho_{eq}$. We recall that
   the {\it basic assumptions of the analysis are Batch Normalization and
   Weight Decay}. Our predictions are for the square loss but we show
 that they should hold also in the case of
 crossnetropy, explored in \cite{Papyan24652}.

 \item At a close to zero loss critical point of  the flow, 
$\nabla_{V_k} f(x_j) = V_k f(x_j)$ with $x_j$ in
   the training set, which are  {\it powerful constraints} on the weight
   matrices to which training converges.  A specific
   dependence of the matrix at each layer on matrices at the other
   layers is thus required. In particular, there are specific relations for each layer matrix
   $V_k$ of the type, explained in the Appendix,

\begin{equation}
  V_k f=[V_L D_{L-1}(x)
  V_{L-1}  \cdots V_{k+1} D_k(x)]^T D_{k-1}(x)
  V_{k-1}D_{k-2}(x)  \cdots D_1(x)  V_1 x,
\end{equation}
\noindent where the $D$ matrices are diagonal with components either
$0$ or $1$, depending on whether the corresponding ReLU unit is on or off. 

As described in \cite{PoggioLiaoArxiv2020} {\it for linear networks, a
  class of possible solutions to these constraint equations are
  projection matrices; another one are orthogonal matrices and more
  generally orthogonal Stiefel matrices on the sphere}. These are {\it
  sufficient but not necessary} conditions to satisfy the constraint
equations. Interestingly, randomly initialized weight matrices (an
extreme case of the NTK regime) are approximately orthogonal.
\end{itemize}

\section{Explicit Regularization and Implicit Dynamic Bias}

We have established here convergence of the gradient flow to a minimum
norm solution for the square loss, when gradient descent is used with
BN (or WN) and WD. This result assumes explicit regularization (WD)
and is thus consistent  with 
\cite{Vardi2020ImplicitRI}, where is is proved that implicit regularization with
the square loss cannot be characterized by any explicit function of
the model parameters. We also show, however, that in the absence of WD
and BN good solutions for {\it binary classification} can be found
because of the {\it  bias in the dynamics of GD toward small
norm solution introduced by near-zero initial conditions}. Again, this
is consistent with \cite{Vardi2020ImplicitRI}, because we identify an
implicit bias in the dynamics which is relevant for classification.

For the exponential loss, BN is strictly not needed since minimization
of the exponential loss maximizes the margin and minimizes the norm
without BN, independently of initial conditions. Thus under the
exponential loss, we expect a margin maximization bias for
$t \to \infty$ as shown in \cite{PNAS2020}, independently of initial
conditions. The effect however can require very long times and
unreasonably high precision to have significant effect in practice.

If there exist several almost-interpolating solutions with the same
norm $\rho_{eq}$, they also have the same margin for each of the
training data.  Though they have the same norm and the same margin on
each of the data point, they may in principle have different ranks of the weight
matrices or of the rank of the local Jacobian
$\frac{\partial f_n}{\partial V_k}$ (at the minimum $W^*$) . Notice
that in deep linear networks the GD dynamics biases the
solution towards small rank solutions, since large eigenvalues
converge much faster the small ones \cite{2019arXiv190912051G}. It in
unclear whether the rank has a role in our analysis of generalization
and we conjecture it does not.

Why does GD have difficulties in converging in the absence of BN+WD,
especially for very deep networks? For square loss, the best answer is
that good tuning of the learning rate is important and BN together
with weight decay was shown to provide a remarkable autotuning
\cite{DBLP:journals/corr/abs-1812-03981}. A related answer is that
regularization is needed to provide stability, including numerical
stability.

\section{Summary}

The main results of the paper analysis can be summarized in the
following

\begin{lemma}
If the gradient flow with 
normalization  and weight decay converges to an
interpolating solution with near-zero square loss, the following
properties hold:

\begin{enumerate}
\item The global minima in the square loss with the smallest $\rho$
  are the global minimum norm
  solutions and have the best margin and the best bound on expected error;
\item Conditions that favour convergence to such minimum norm
  solutions are batch normalization with weight decay ($\lambda>0$)
  and small initialization (small $\rho$);
  \item initialization with small $\rho(0)$ can be sufficient to
    induce a bias in the dynamics of GD leading to large margin
    minimizers of the square loss;
\item The condition
  $\frac{\partial f(x_j)}{\partial V_k} = V_k f(x_j)$ which holds at
  the critical points of the SGD dynamics that are global minima, is key
  in predicting several properties of the {\it Neural
    Collapse\cite{Papyan24652}};
    \item the same condition represents a powerful constraint on the
      set of weight matrices at
      convergence.
\end{enumerate}
\end{lemma}

\subsection{Discussion}

  The role of the Lagrange multiplier term $\nu \sum_k
    ||V_k||^2 $ in Equation \ref{LagrangeMultiplierEquation} is
    different from a standard regularization term because $\nu$,
    determined by the constraint $||V_k||=1$ can be positive or
    negative, depending on the sign of the error $\nu=-
    \sum_n(\rho^2 f_n^2-\rho y_nf_n)$. Thus the $\nu$ term acts as
    a regularizer when the norm of $V_k$ is larger than $1$ but has
    the opposite effect for $||V_k|| <1$, thus constraing each $V_k$
    to the unit sphere. For the exponential loss, the situation is
    different and $\nu$ in  $\nu \sum_k
    ||V_k||^2 $ acts as a positive regularization parameter, albeit a
    vanishing one (for $t \to \infty$).

   Are there any implications of the theory sketched here for
    mechanisms of learning in cortex? Somewhat intriguingly, some form
    of normalization, often described as a balance of excitation and inhibition, has long
    been thought to be a key function of intracortical circuits
    in cortical areas\cite{DouglasMartin2004}. One of the first deep models of visual cortex
    models, HMAx, explored the biological plausibility of specific
    normalization circuits with spiking and non-spiking neurons. It is
    also interesting to note that the Oja rule describing synaptic
    plasticity in terms of changes to the synaptic weight
    is the Hebb rule plus a normalization term that corresponds to a
    Lagrange multiplier.

  The main problems left open by this paper are:
    \begin{itemize}

      \item The analysis is so far restricted to gradient flow. It
        should be exteded to  {\it gradient descent} along the lines
        of \cite{DBLP:journals/corr/abs-1812-03981}.
        \item It is remarkable that for the case of no BN
          and no WD, the  dynamical system still yields good results, provided
         {\it initialization is small}. The case of BN+WD is the only one
          which seems rather independent of initial conditions in our experiments.
          \item In this context, an extension of the analysis to SGD may also be critical for
  providing a satisfactory analysis of convergence.
 \end{itemize}

	\vskip0.1in

        {\bf Acknowledgments} We are grateful to Shai Shalev-Schwartz,
        Andrzej Banbuski, Arturo Desza, Akshay Rangamani, Santosh
        Vempala, David Donoho, Vardan Papyan, X.Y. Han, Silvia Villa and especially
        to Eran Malach for very useful comments.  This material is
        based upon work supported by the Center for Minds, Brains and
        Machines (CBMM), funded by NSF STC award CCF-1231216, and part
        by C-BRIC, one of six centers in JUMP, a Semiconductor
        Research Corporation (SRC) program sponsored by DARPA.

	\newpage

	\small
	\bibliographystyle{unsrt}
	\bibliography{Boolean}

\begin{thebibliography}{10}

\bibitem{2019arXiv190507325S}
Mor {Shpigel Nacson}, Suriya {Gunasekar}, Jason~D. {Lee}, Nathan {Srebro}, and
  Daniel {Soudry}.
\newblock {Lexicographic and Depth-Sensitive Margins in Homogeneous and
  Non-Homogeneous Deep Models}.
\newblock {\em arXiv e-prints}, page arXiv:1905.07325, May 2019.

\bibitem{DBLP:journals/corr/abs-1906-05890}
Kaifeng Lyu and Jian Li.
\newblock Gradient descent maximizes the margin of homogeneous neural networks.
\newblock {\em CoRR}, abs/1906.05890, 2019.

\bibitem{theory_III}
A.~Banburski, Q.~Liao, B.~Miranda, T.~Poggio, L.~Rosasco, B.~Liang, and
  J.~Hidary.
\newblock Theory of deep learning \uppercase{III}: Dynamics and generalization
  in deep networks.
\newblock {\em CBMM Memo No. 090}, 2019.

\bibitem{hui2020evaluation}
Like Hui and Mikhail Belkin.
\newblock Evaluation of neural architectures trained with square loss vs
  cross-entropy in classification tasks, 2020.

\bibitem{PoggioLiaoArxiv2020}
T.~Poggio and Q.~Liao.
\newblock Generalization in deep network classifiers trained with the square
  loss.
\newblock {\em CBMM Memo No. 112}, 2019.

\bibitem{ioffe2015batch}
Sergey Ioffe and Christian Szegedy.
\newblock Batch normalization: Accelerating deep network training by reducing
  internal covariate shift.
\newblock {\em arXiv preprint arXiv:1502.03167}, 2015.

\bibitem{SalDied16}
Tim Salimans and Diederik~P. Kingm.
\newblock Weight normalization: A simple reparameterization to accelerate
  training of deep neural networks.
\newblock {\em Advances in Neural Information Processing Systems}, 2016.

\bibitem{DBLP:journals/corr/abs-1812-03981}
Sanjeev Arora, Zhiyuan Li, and Kaifeng Lyu.
\newblock Theoretical analysis of auto rate-tuning by batch normalization.
\newblock {\em CoRR}, abs/1812.03981, 2018.

\bibitem{PoggioCooper}
T.~Poggio and Y.~Cooper.
\newblock Loss landscape: Sgd has a better view.
\newblock {\em CBMM Memo 107}, 2020.

\bibitem{PoggioCooper2020}
T.~Poggio and Y.~Cooper.
\newblock Loss landscape: Sgd can have a better view than gd.
\newblock {\em CBMM memo 107}, 2020.

\bibitem{Foundations}
Tomaso Poggio.
\newblock Stable foundations for learning.
\newblock {\em Center for Brains, Minds and Machines (CBMM) Memo No. 103},
  2020.

\bibitem{Papyan24652}
Vardan Papyan, X.~Y. Han, and David~L. Donoho.
\newblock Prevalence of neural collapse during the terminal phase of deep
  learning training.
\newblock {\em Proceedings of the National Academy of Sciences},
  117(40):24652--24663, 2020.

\bibitem{Vardi2020ImplicitRI}
Gal Vardi and O.~Shamir.
\newblock Implicit regularization in relu networks with the square loss.
\newblock 2020.

\bibitem{PNAS2020}
Tomaso Poggio, Andrzej Banburski, and Qianli Liao.
\newblock Theoretical issues in deep networks.
\newblock {\em PNAS}, 2020.

\bibitem{2019arXiv190912051G}
Daniel {Gissin}, Shai {Shalev-Shwartz}, and Amit {Daniely}.
\newblock {The Implicit Bias of Depth: How Incremental Learning Drives
  Generalization}.
\newblock {\em arXiv e-prints}, page arXiv:1909.12051, September 2019.

\bibitem{DouglasMartin2004}
RJ~Douglas and KA~Martin.
\newblock Neuronal circuits of the neocortex.
\newblock {\em Annu Rev Neuroscience}, 27:419--51, 2004.

\end{thebibliography}
	\normalsize

\begin{figure}
  \centering
  \includegraphics[width=1.0\textwidth]{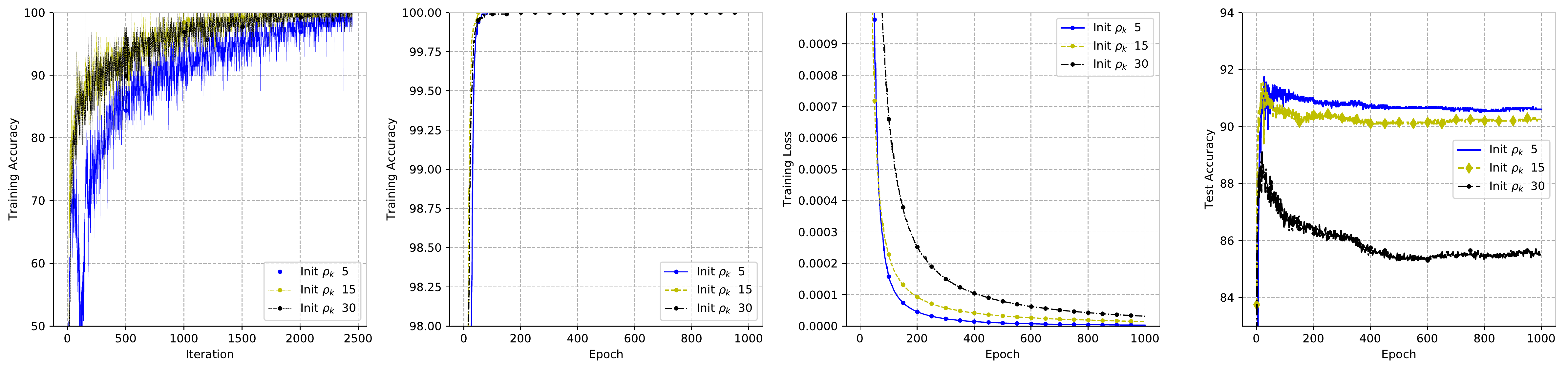}   
  \caption {\it \textbf{ConvNet, no Batch Normalization, no Weight
      Decay}. Binary classification on two classes from CIFAR-10,
    trained with MSE loss. The model is a very simple network with 4
    layers of fully-connected Layers. The ReLU nonlinearity is used. The weight matrices of all layers are initialized with zero-mean normal distribution, scaled by a constant such that the Frobenius norm of each matrix is either 5, 15 or 30. We run SGD with batch size 128, constant learning rate 0.1 and momentum 0.9 for 1000 epochs..  No data augmentation.    Every input to the network is scaled such that it has Frobenius norm 1.  }             
  \label{fig:training_val}
\end{figure}

\begin{figure}
  \centering
  \includegraphics[width=1.0\textwidth]{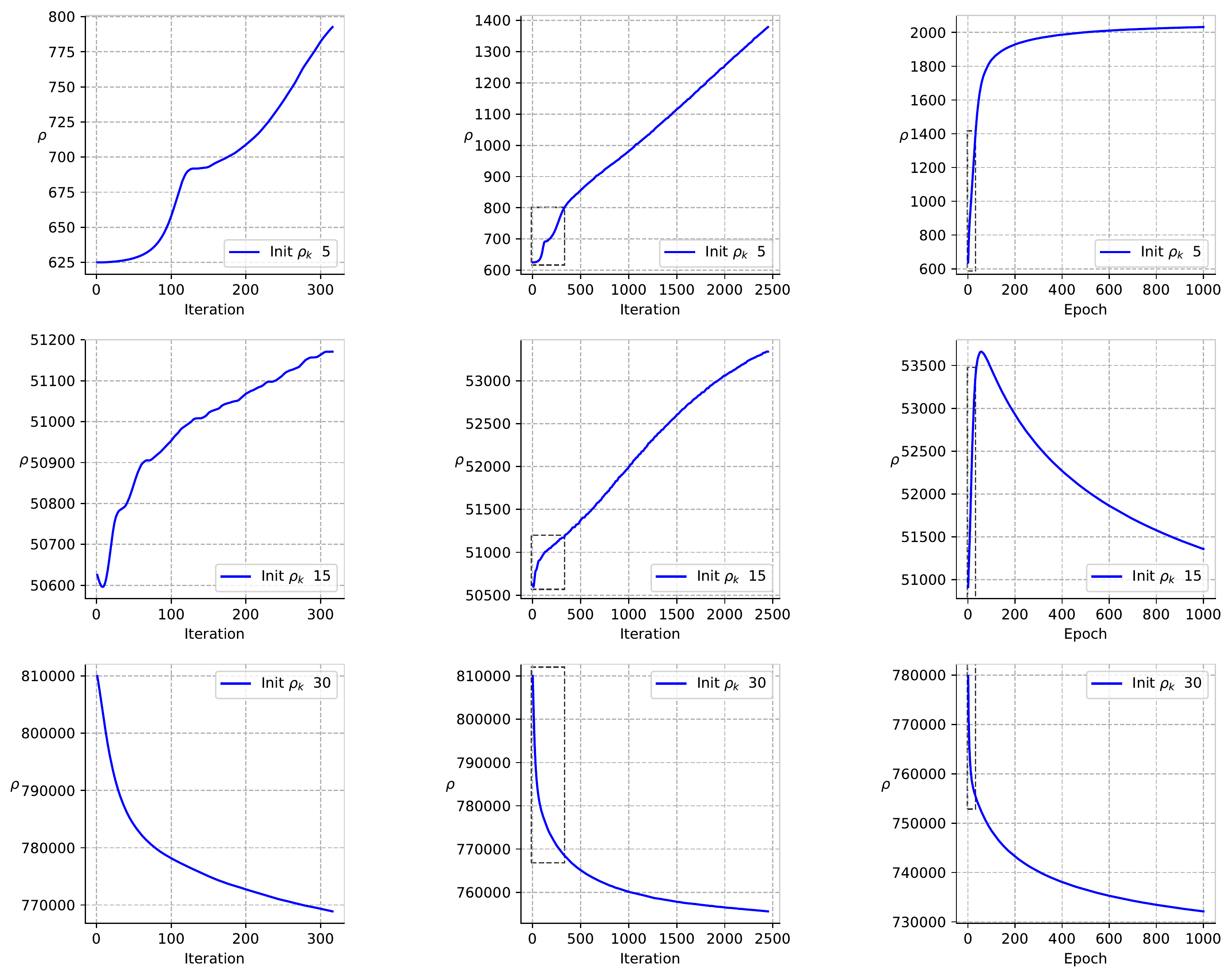}
  \caption {\it \textbf{ConvNet, no Batch Normalization, no Weight
      Decay}. Dynamics of $\rho$ from experiments in Figure \ref{fig:training_val}. First row: small initialization (5).  Second row: large initialization (15).  Third row: extra large initialization (30). A dashed rectangle denotes the previous subplot's domain and range in the new subplot.    More details to be added.  }      
  \label{fig:rho}
\end{figure}

\begin{figure}
  \centering
  \includegraphics[width=1.0\textwidth]{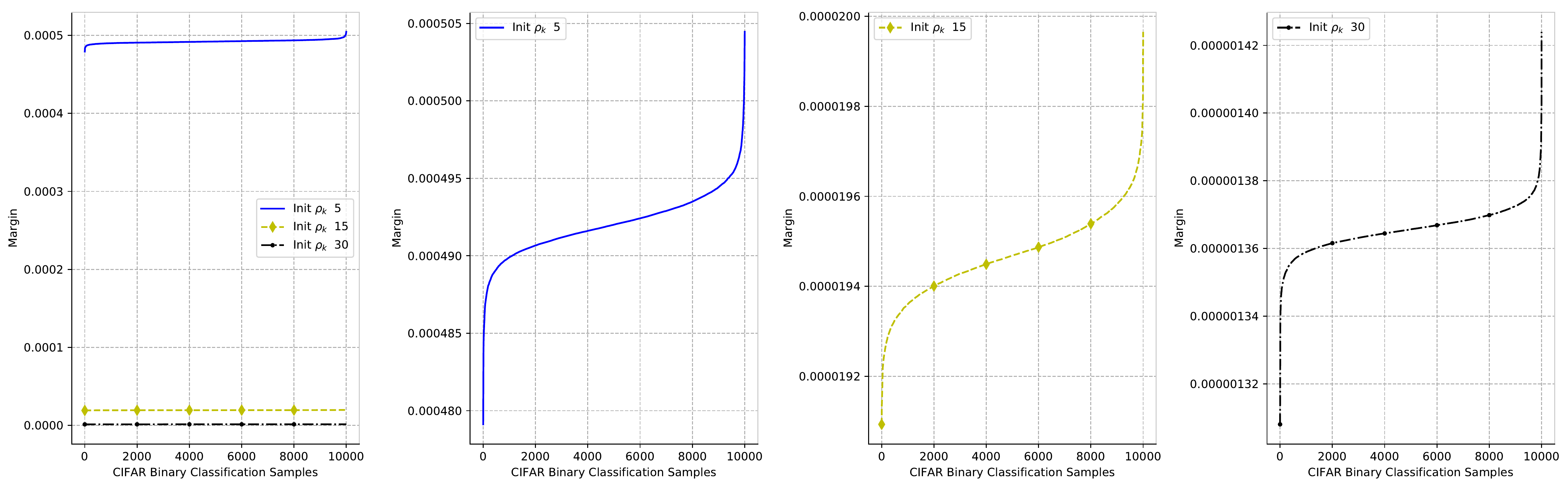} 
  \caption {\it \textbf{ConvNet, no Batch Normalization, no Weight
      Decay}. Margin of all training samples}  \label{fig:margin}
\end{figure}

\end{document}